\documentclass[11pt]{article}
\usepackage{paclic35}
\usepackage{times}
\usepackage{multirow}
\usepackage{textcomp}
\usepackage[T1]{fontenc}
\usepackage[T5]{fontenc}
\usepackage[utf8]{inputenc}
\usepackage{amssymb,amsfonts}
\usepackage{colortbl, hhline, tabularx}
\usepackage{booktabs}
\usepackage{graphicx}
\usepackage{fdsymbol}
\usepackage{IEEEtrantools}
\usepackage[mathscr]{euscript}
\usepackage{color}
\usepackage{xurl}
\usepackage{calc}
\usepackage{placeins}
\usepackage{float}
\usepackage{makecell}
\usepackage{siunitx}
\usepackage{latexsym}
\usepackage{etoolbox}
\usepackage{algorithm,algpseudocode}

\protect\robustify\bfseries

\DeclareMathOperator*{\argmax}{argmax}

\newcommand{\dgr}{\raise0.65ex\hbox{\tiny$\dagger$}}
\newcommand{\ddgr}{\raise0.65ex\hbox{\tiny$\ddagger$}}

\newcolumntype{j}{>{\columncolor[gray]{0.99}}S[table-format=1.2,table-space-text-pre=~~,,table-space-text-post=\ddgr]}
\newcolumntype{z}{>{\columncolor[gray]{0.97}}r}

\newcolumntype{x}{>{\columncolor[gray]{0.97}}r}
\newcolumntype{g}{>{\columncolor[gray]{0.97}}S[table-format=1.2,table-space-text-pre=~~,table-space-text-post=\ddgr]}
\newcolumntype{k}{>{\columncolor[gray]{0.93}}S[table-format=1.2,table-space-text-pre=~~,table-space-text-post=\ddgr]}

\title{Joint Chinese Word Segmentation and Part-of-speech Tagging \\ via Two-stage Span Labeling}

\author{Duc-Vu Nguyen\textsuperscript{1,3}, Linh-Bao Vo\textsuperscript{2,3}, Ngoc-Linh Tran\textsuperscript{2,3}\\\textbf{Kiet Van Nguyen\textsuperscript{2,3}, Ngan Luu-Thuy Nguyen\textsuperscript{2,3}}\\\textsuperscript{1}Multimedia Communications Laboratory, University of Information Technology,\\Ho Chi Minh City, Vietnam \\
\textsuperscript{2}University of Information Technology, Ho Chi Minh City, Vietnam\\
\textsuperscript{3}Vietnam National University, Ho Chi Minh City, Vietnam\\\tt{vund@uit.edu.vn, \{18520503,20521538\}@gm.uit.edu.vn}\\ \tt{\{kietnv,ngannlt\}@uit.edu.vn}}

\date{}

\begin{document}
\maketitle
\begin{abstract}
Chinese word segmentation and part-of-speech tagging are necessary tasks in terms of computational linguistics and application of natural language processing.
Many researchers still debate the demand for Chinese word segmentation and part-of-speech tagging in the deep learning era.
Nevertheless, resolving ambiguities and detecting unknown words are challenging problems in this field.
Previous studies on joint Chinese word segmentation and part-of-speech tagging mainly follow the character-based tagging model focusing on modeling n-gram features.
Unlike previous works, we propose a neural model named \textsc{SpanSegTag} for joint Chinese word segmentation and part-of-speech tagging following the span labeling in which the probability of each n-gram being the word and the part-of-speech tag is the main problem.
We use the biaffine operation over the left and right boundary representations of consecutive characters to model the n-grams.
Our experiments show that our BERT-based model \textsc{SpanSegTag} achieved competitive performances on the CTB5, CTB6, and UD, or significant improvements on CTB7 and CTB9 benchmark datasets compared with the current state-of-the-art method using BERT or ZEN encoders.
\end{abstract}

\section{Introduction}
Chinese word segmentation (CWS) and part-of-speech (POS) tagging are necessary tasks in terms of computational linguistics and application of natural language processing (NLP). There are two primary approaches for joint CWS and POS tagging, including the two-step and one-step methods. The two-step approach is to find words and then assign POS tags to found words. \newcite{ng-low-2004-chinese} proposed the one-step approach that combines CWS and POS tagging into a unified joint task. The one-step approach was proved better than two-step approach by many prior studies \cite{jiang-etal-2008-cascaded,jiang-etal-2009-automatic,sun-2011-stacked,zeng-etal-2013-graph,zheng-etal-2013-deep,kurita-etal-2017-neural,shao-etal-2017-character,Zhang2018ASA}. These studies proposed various methods incorporating linguistic features or contextual information into their joint model. Remarkably, \newcite{tian-etal-2020-joint} proposed a two-way attention mechanism incorporating both context features and corresponding syntactic knowledge from off-the-shelf toolkits for each input character.

\begin{figure}[!ht]
\centering
\includegraphics[width=0.9\columnwidth]{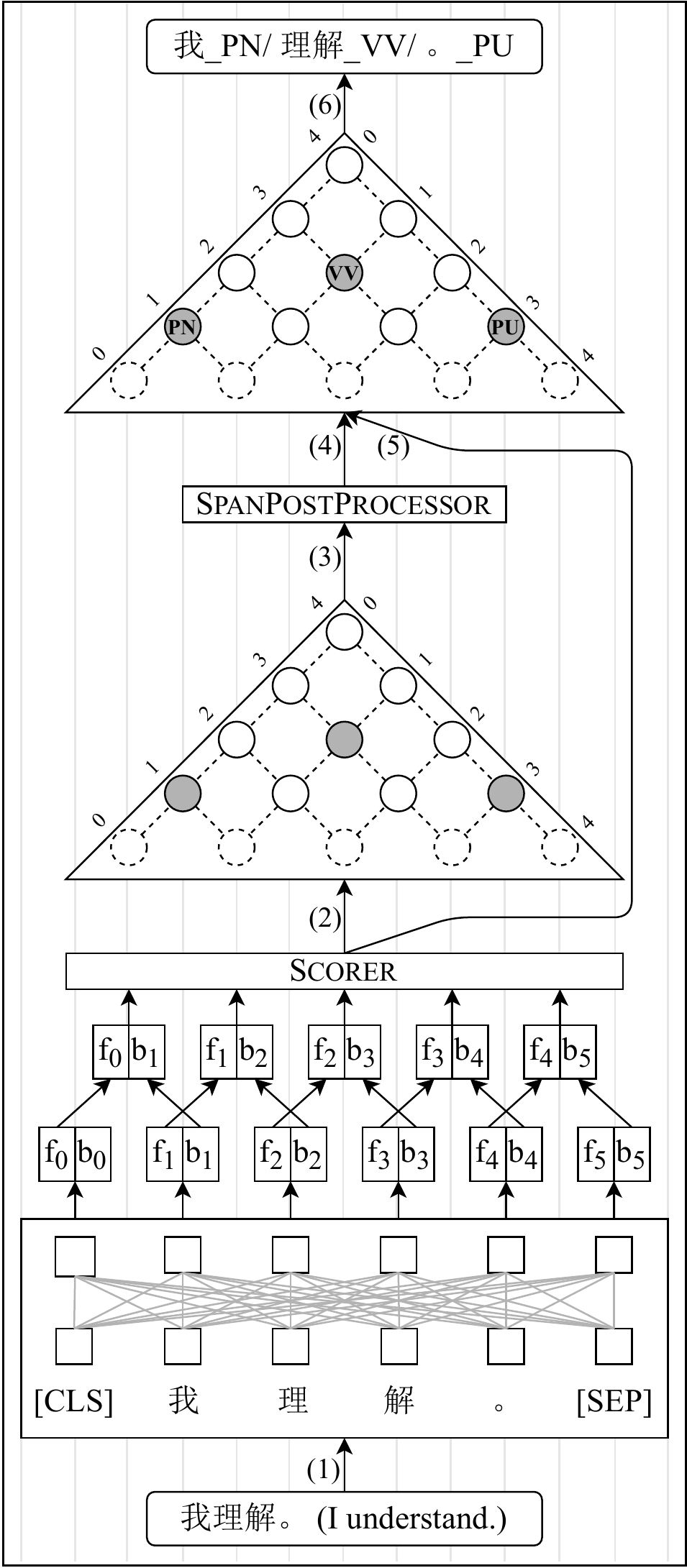}
\caption{\label{architecture}The architecture of \textsc{SpanSegTag} for the joint CWS and POS tagging with two stages via span labeling: word segmentation and POS tagging.}
\end{figure}

To our best knowledge, we observed all previous studies for joint CWS and POS tagging following the character-based tagging paradigm. The character-based tagging effectively produces the best combination of word boundary and POS tag. However, this character-based tagging paradigm does not give us a clear explanation when processing overlapping ambiguous strings. From the view of experimental psychology, human perception and performance, \newcite{Ma_2014} concluded that multiple words constituted by the characters in the perceptual span are activated when processing overlapping ambiguous strings. Besides, \newcite{tian-etal-2020-improving-chinese} shown that modeling word-hood for n-gram information is essential for CWS. Next, the current state-of-the-art method for joint CWS and POS tagging also confirmed the importance of modeling words and their knowledge, e.g., POS tag \cite{tian-etal-2020-joint}.

The previous studies in two views of experimental psychology, human perception and performance, \cite{Ma_2014} and computational linguistics \cite{tian-etal-2020-improving-chinese,tian-etal-2020-joint} inspired us to propose the span labeling approach for joint CWS and POS tagging. To avoid the model size dependent on numbers of n-grams and their corresponding POS tag, we use span to model n-gram and n-gram with POS tag instead of using the memory networks in \cite{tian-etal-2020-improving-chinese,tian-etal-2020-joint}. More particularly, inspired by \newcite{stern-etal-2017-minimal}, \newcite{ijcai2020-560}, and \cite{vund-etal-2021-spanseg}, we use the biaffine operation over the left and right boundary representations of consecutive characters to model n-grams and their POS tag. As the prior work of \newcite{vund-etal-2021-spanseg}, we use a simple post-processing heuristic algorithm instead of using other models to deal with the overlapping ambiguity phenomenon \cite{li-etal-2003-unsupervised,gao-etal-2005-chinese}. Finally, we experimented with BiLSTM \cite{hochreiter97} and BERT encoders \cite{devlin-etal-2019-bert}.

Our experiments show that our BERT-based model \textsc{SpanSegTag} achieved competitive performances on the CTB5, CTB6, UD1, and UD2, and significant improvements on the two large benchmark datasets CTB7 and CTB9 compared with the current state-of-the-art method using BERT or ZEN encoders \cite{tian-etal-2020-joint}. Our \textsc{SpanSegTag} did not perfectly perform in five Chinese benchmark datasets. However, \textsc{SpanSegTag} achieved a good recall of in-vocabulary words and their POS tag scores on CTB6, CTB7 and CTB9 datasets. This score is used to measure the performance of the segmenter in resolving ambiguities in word segmentation \cite{gao-etal-2005-chinese}.

\section{The Proposed Framework}
We present the architecture of our proposed framework, namely \textsc{SpanSegTag}, for joint CWS and POS tagging in Figure~\ref{architecture}. As we can see in Figure~\ref{architecture}, data path (1) indicates the input sentence to be fed into the BERT encoder. The hidden state vector from the BERT encoder is chunk into two vectors with the same size as the forward and backward vectors in the familiar encoder, BiLSTM. Next, all boundary representations are fed into the \textsc{Scorer} module. Data path (2) indicates the span representations for the word segmentation task, and data path (5) indicates the span representations for the POS tagging task. Data path (3) indicates predicted spans representing predicted word boundaries. The \textsc{SpanPostProcessor} module produces the predicted spans satisfying non-overlapping between every two spans. Finally, given data paths (4) and (5), the data path (6) indicates the joint CWS and POS tagging.


\subsection{Joint Chinese Word Segmentation and Part-of-speech Tagging as Two Stages Span Labeling}
\label{formalization}

The input sentence of joint CWS and POS tagging is a sequence of characters $\mathcal{X} = x_\text{1} x_\text{2} \dots x_{n}$ with the length of $n$. Given the input sentence $\mathcal{X}$, the output of CWS is a sequence of words $\mathcal{W} = w_\text{1} w_\text{2} \dots w_{m}$ with the length of $m$, and the output of Chinese POS tagging is a sequence of POS tags $\mathcal{T} = t_\text{1} t_\text{2} \dots t_{m}$ with the length of $m$, where $\text{1} \leq m \leq n$. Besides, we have a property that the Chinese word $w_j$ is constituted by one Chinese character or consecutive characters. Therefore, we use the sequence of characters $x_i x_{i+\text{1}} \dots x_{i+k-\text{1}}$ to denote that the word $w_j$ is constituted by $k$ consecutive characters beginning at character $x_i$, where $\text{1} \leq k \leq n$ and $k = \text{1}$ representing single words and $\text{2} \leq k \leq n$ representing compound words. We get the inspiration of span representation in constituency parsing \cite{stern-etal-2017-minimal} to use the span $(i - \text{1}, i - \text{1} + k)$ representing the word constituted by $k$ consecutive characters $x_i x_{i+\text{1}} \dots x_{i+k-\text{1}}$ beginning at character $x_i$, where $i - \text{1}$ and $i - \text{1} + k$ are the left and the right boundary index of word $x_i x_{i+\text{1}} \dots x_{i+k-\text{1}}$, respectively.

After presenting notations, we propose our approach for the joint CWS and POS tagging problem. Firstly, to our knowledge, most recent works focus on modeling the probability that a Chinese character can be one in the combination of $\{\text{B, I, E, S}\}$ and Chinese POS tags set. Next, the current state-of-the-art method for CWS approaching BIES tagging of \newcite{tian-etal-2020-improving-chinese} proposed word-hood memory to model n-gram information. Additionally, the current state-of-the-art method for joint CWS and POS tagging approaching BIES tagging of \newcite{tian-etal-2020-joint} shown that modeling n-gram knowledge, e.g., word and POS tag, is essential. Therefore, we get inspiration of \newcite{tian-etal-2020-improving-chinese} and \newcite{tian-etal-2020-joint} to focus on modeling words and POS tags in a straightforward way rather than modeling BIES tags of characters. Given the input sentence $\mathcal{X}$, our idea is to model the probability that the consecutive Chinese characters can be a word via one formulation. Similarly, given the input sentence $\mathcal{X}$, we also model the probability that consecutive Chinese characters can be assigned a specific POS tag or the non-word tag via one formulation. To summarize, given span representations, we formalize the joint CWS and POS tagging task as two continuous sub-tasks in our \textsc{SpanSegTag} as following: (i) binary classification dealing with word segmentation; (ii) multi-class classification dealing with POS tagging.

Formally, the first stage of our \textsc{SpanSegTag} for CWS can be formalized as:
\begin{align}
\hat{\mathcal{S}}_{\text{novlp}} = \textsc{SpanPostProcessor}(\hat{\mathcal{S}})\label{eq:formal}
\end{align}
where \textsc{SpanPostProcessor}($\hat{\mathcal{S}}$) is introduced in the work of \newcite{vund-etal-2021-spanseg}. \textsc{SpanPostProcessor}($\hat{\mathcal{S}}$) solely is an algorithm for producing the word segmentation boundary guaranteeing non-overlapping between every two spans. The $\hat{\mathcal{S}}$ is the set of predicted spans as follows:
\begin{align}
\hat{\mathcal{S}} = \bigg\{(l, r)~\text{for}~\text{0} \leq l \leq n - \text{1} ~\text{and}~l < r \leq n\nonumber\\
\text{and}~\textsc{Scorer}(\mathcal{X}, l, r).\textsc{Seg} > \text{0.5}\bigg\}\label{eq:seg}
\end{align}
where $n$ is the length of the input sentence. The $l$ and $r$ denote left and right boundary indexes of the specific span. The \textsc{Scorer}($\mathcal{X}, l, r$).\textsc{Seg} is the scoring module for the span $(l, r)$ of sentence $\mathcal{X}$. The output of \textsc{Scorer}($\mathcal{X}, l, r$).\textsc{Seg} has a value in the range of 0 to 1. We choose the sigmoid function as the activation function at the last layer of \textsc{Scorer}($\mathcal{X}, l, r$).\textsc{Seg} module.

Next, given the set of predicted spans $\hat{\mathcal{S}}_{\text{novlp}}$ satisfying non-overlapping between every two spans for the input sentence $\mathcal{X}$, the second stage of our \textsc{SpanSegTag} to perform Chinese POS tagging can be formalized as:
\begin{align}
\hat{\mathcal{Y}} = \bigg\{\Big((l,r),\argmax_{\hat{t} \in \mathcal{T}} \textsc{Scorer}(\mathcal{X}, l, r).\textsc{Tag}[\hat{t}]\Big)\nonumber\\
\text{for}~(l,r) \in \hat{\mathcal{S}}_{\text{novlp}} \bigg\}\label{eq:tag}
\end{align}
where $\mathcal{T}$ is the union of Chinese POS tag set and the non-word tag since the $\hat{\mathcal{S}}_{\text{novlp}}$ can include the incorrectly predicted span. The \textsc{Scorer}($\mathcal{X}, l, r$).\textsc{Tag}$[\hat{t}]$ is the scoring module for the span $(l, r)$ of sentence $\mathcal{X}$ assigned tag $\hat{t}$. To sum up, given the input sentence $\mathcal{X}$, the set $\hat{\mathcal{Y}}$ includes predicted spans with the POS tag. Therefore, the set $\hat{\mathcal{Y}}$ is the result of the second stage of our \textsc{SpanSegTag} and of the joint CWS and POS tagging task.

The main idea of our \textsc{SpanSegTag} is formalized through three Equations~\ref{eq:formal},~\ref{eq:seg}, and~\ref{eq:tag}. To train our \textsc{SpanSegTag}, we have to optimize parameters in \textsc{Scorer}($\mathcal{X}, l, r$).\textsc{Seg} and \textsc{Scorer}($\mathcal{X}, l, r$).\textsc{Tag}$[\hat{t}]$ modules. As we clearly see that there is no parameters in the \textsc{SpanPostProcessor}($\hat{\mathcal{S}}$) module. However, the optimization of parameters in \textsc{Scorer}($\mathcal{X}, l, r$).\textsc{Tag}$[\hat{t}]$ based on the $\hat{\mathcal{S}}_{\text{novlp}}$ indirectly optimizes parameters in our \textsc{SpanSegTag} by learning from the result of \textsc{SpanPostProcessor}($\hat{\mathcal{S}}$). For example, if an incorrect span is assigned non-word tag, then our \textsc{SpanSegTag} is trained to deal with this case via \textsc{Scorer}($\mathcal{X}, l, r$).\textsc{Tag}$[\hat{t}]$ module.

Therefore, the cost function for training our \textsc{SpanSegTag} is the combined loss of binary classification and multi-class classification. The cost function for training CWS in our \textsc{SpanSegTag} is
\begin{align}
& J_{\textsc{Seg}}(\theta,\theta_{\textsc{Seg}}) = -\frac{1}{|\mathcal{D}|}\sum_{\mathcal{X, S} \in \mathcal{D}} \Bigg( & \nonumber\\
& \frac{1}{\big(n(n+1)\big)/2} \sum_{l=0}^{n-1} \sum_{r=l+1}^{n} \bigg( & \nonumber\\
& \left[(l,r) \in \mathcal{S}\right]\log\Big(\textsc{Scorer}(\mathcal{X}, l, r).\textsc{Seg}\Big) & \nonumber\\
+ & \left[(l,r) \notin \mathcal{S}\right]\log\Big(1 - \textsc{Scorer}(\mathcal{X}, l, r).\textsc{Seg}\Big)\bigg) \Bigg) &
\end{align}
where $\mathcal{D}$ is the training set and $|\mathcal{D}|$ is the size of the training set. For each pair ($\mathcal{X, S}$) in training set $\mathcal{D}$, we compute binary cross-entropy loss for all spans $(l, r)$, where $\text{0} \leq l \leq n - \text{1} ~\text{and}~l < r \leq n$, and $n$ is the length of sentence $\mathcal{X}$. The term $\left[(l,r) \in \mathcal{S}\right]$ has the value of 1 if span $(l, r)$ belongs to the list $\mathcal{S}$ of sentence $\mathcal{X}$ and conversely, of 0. Similarly, the term $\left[(l,r) \notin \mathcal{S}\right]$ has the value of 1 if span $(l, r)$ does not belong to the list $\mathcal{S}$ of sentence $\mathcal{X}$ and conversely, of 0. Notably, our training and prediction progress, we discard spans with length greater than 7 as the maximum n-gram length following \cite{diao-etal-2020-zen} to reduce negative spans.

Next, the cost function for training Chinese POS tagging in our \textsc{SpanSegTag} is the cross entropy loss:
\begin{align}
& J_{\textsc{Tag}}(\theta,\theta_{\textsc{Tag}}) = \frac{1}{|\mathcal{D}|}\sum_{\mathcal{X, Y} \in \mathcal{D}} \Bigg(\frac{1}{|\hat{\mathcal{S}}_{\text{novlp}}|}\sum_{(l,r) \in \hat{\mathcal{S}}_{\text{novlp}}}\nonumber\\
& \bigg(-\textsc{Scorer}(\mathcal{X}, l, r).\textsc{Tag}[t]\nonumber\\
& + \log\Big(\sum_{\hat{t} \in \mathcal{T}}\exp\big(\textsc{Scorer}(\mathcal{X}, l, r).\textsc{Tag}[\hat{t}]\big)\Big) \bigg) \Bigg)
\end{align}
where $t$ denotes the truth label of span $(l, r)$ from $\mathcal{Y}$ in the input sentence $\mathcal{X}$. Finally, the cost function for training our \textsc{SpanSegTag} is 
\begin{align}
J(\theta, \theta_\textsc{Seg}, \theta_\textsc{Tag}) = J_{\textsc{Seg}}(\theta,\theta_{\textsc{Seg}}) + J_{\textsc{Tag}}(\theta,\theta_{\textsc{Tag}})
\end{align}

\subsection{Decoding Algorithm for Predicted Span}
\label{decoding}

As the problem in prior work of \newcite{vund-etal-2021-spanseg}, in the predicted span set $\hat{\mathcal{S}}$ mentioned in Equation~\ref{eq:seg} there exists overlapping between some two spans. To solve this, \newcite{vund-etal-2021-spanseg} keep the spans with the highest score and eliminate the remainder. The overlapping ambiguity phenomenon happens during our \textsc{SpanSegTag} predicting compound words. Additionally, our \textsc{SpanSegTag} encounters the missing word boundary problem. That problem can be caused by originally predicted spans, the consequence of solving overlapping ambiguity, or more than seven-character spans mentioned in subsection~\ref{formalization}. Finally, we add the missing word boundary based on all predicted spans $(i - \text{1}, i - \text{1} + k)$ with $k = \text{1}$ to single words to deal with the missing word boundary problem following \newcite{vund-etal-2021-spanseg}. The detail of this algorithm is shown in the work of \newcite{vund-etal-2021-spanseg}. To sum up, \textsc{SpanPostProcessor}($\hat{\mathcal{S}}$) is considered as the heuristic algorithm, while the inference algorithm in \cite{ye-ling-2018-hybrid} is optimal.

\subsection{Span Scoring}
\label{scoring}
Inspired by \newcite{ijcai2020-560}, the span scoring module $\textsc{Scorer}(\mathcal{X}, l, r).\textsc{Seg}$ for finding probability of word is computed by using a biaffine operation over the left boundary representation of character $x_{l}$ and the right boundary representation of character $x_{r}$:
\begin{align}
&\textsc{Scorer}(\mathcal{X}, l, r).\textsc{Seg} =\text{sigmoid}\bigg( \nonumber \\
&\begin{bmatrix}\text{MLP}_\text{seg}^{\text{left}}(\textbf{f}_{l}\oplus \textbf{b}_{l+\text{1}}) \\ \text{1}\end{bmatrix}^{\text{T}}\textbf{W}\big(\text{MLP}_\text{seg}^{\text{right}}(\textbf{f}_{r}\oplus \textbf{b}_{r+\text{1}})\big)\bigg)
\end{align}
where $\textbf{W} \in \mathbb{R}^{(d+\text{1})\times d}$ and the symbol $\oplus$ denote the concatenation operation. Similarly, the span scoring module $\textsc{Scorer}(\mathcal{X}, l, r).\textsc{Tag}[\hat{t}]$ for finding score of a POS tag $\hat{t} \in \mathcal{T}$ is computed by:
\begin{align}
&\textsc{Scorer}(\mathcal{X}, l, r).\textsc{Tag}[\hat{t}] = \nonumber \\
&\begin{bmatrix}\text{MLP}_\text{tag}^{\text{left}}(\textbf{f}_{l}\oplus \textbf{b}_{l+\text{1}}) \\ \text{1}\end{bmatrix}^{\text{T}}\textbf{W}_{\hat{t}}\begin{bmatrix}\text{MLP}_\text{tag}^{\text{right}}(\textbf{f}_{r}\oplus \textbf{b}_{r+\text{1}}) \\ \text{1}\end{bmatrix}
\end{align}
where $\textbf{W}_{\hat{t}} \in \mathbb{R}^{(d+\text{1})\times (d+\text{1})}$. As mentioned in subsection~\ref{formalization}, we have $\text{0} \leq l \leq n - \text{1} ~\text{and}~l < r \leq n$, where $n$ is the length of input sentence $\mathcal{X}$. The $\text{MLP}_\text{seg}^{\text{left}}$, $\text{MLP}_\text{seg}^{\text{right}}$, $\text{MLP}_\text{tag}^{\text{left}}$ and $\text{MLP}_\text{tag}^{\text{right}}$ are multilayer perceptrons for transforming hidden states from encoder to left and boundary representations with the output dimension of $d$ for CSW and POS tagging tasks. Vectors $\textbf{f}_{i}$ and $\textbf{b}_{i}$ denote forward and backward hidden state vectors from BiLSTM encoder. In case we use BERT encoder, we chunk the hidden state vector from BERT encoder into two vectors with the same size as the  forward and backward hidden state vectors in the BiLSTM encoder.

\subsection{Encoder Architecture}
\label{encoder}
To experiment with our proposed \textsc{SpanSegTag}, we use BiLSTM encoder \cite{hochreiter97} and $\text{BERT}_\text{BASE}$ encoder for Chinese \cite{devlin-etal-2019-bert}. In case we use LSTM encoder, we use character pre-trained Chinese embedding with the dimension of 64 provided \newcite{shao-etal-2017-character}. In case we use BERT encoder, we use only the hidden state of the last layer of BERT as \newcite{tian-etal-2020-joint}.

\section{Experiments}

\subsection{Datasets}

{
\setlength{\tabcolsep}{4pt}
\begin{table}[ht]
\centering
\resizebox{\columnwidth}{!}{%
\begin{tabular}{lczrzc}
\toprule
\multicolumn{2}{c}{\textbf{Datasets}} & \multicolumn{1}{z}{\textbf{\# Sent}} & \multicolumn{1}{c}{\textbf{\# Char}} & \multicolumn{1}{z}{\textbf{\# Word}} & \textbf{OOV} \\
\midrule
\multicolumn{1}{c}{\multirow{3}{*}{CTB5}} & Train & 18,104 & 804,587 & 493,930 & - \\
\multicolumn{1}{c}{} & Dev & 352 & 11,543 & 6,821 & 8.1 \\
\multicolumn{1}{c}{} & Test & 348 & 13,738 & 8,008 & 3.5 \\\midrule
\multicolumn{1}{c}{\multirow{3}{*}{CTB6}} & Train & 23,420 & 1,055,583 & 641,368 & - \\
\multicolumn{1}{c}{} & Dev & 2,079 & 100,316 & 59,955 & 5.4 \\
\multicolumn{1}{c}{} & Test & 2,796 & 134,149 & 81,578 & 5.6 \\\midrule
\multirow{3}{*}{CTB7} & Train & 31,112 & 1,160,209 & 717,874 & - \\
 & Dev & 10,043 & 387,209 & 236,590 & 5.5 \\
 & Test & 10,292 & 398,626 & 245,011 & 5.2 \\\midrule
\multirow{3}{*}{CTB9} & Train & 105,971 & 2,642,998 & 1,696,340 & - \\
 & Dev & 9,850 & 209,739 & 136,468 & 2.9 \\
 & Test & 15,929 & 378,502 & 242,317 & 3.1 \\\midrule
\multirow{3}{*}{UD} & Train & 3,997 & 156,309 & 98,608 & - \\
 & Dev & 500 & 20,000 & 12,663 & 12.1 \\
 & Test & 500 & 19,206 & 12,012 & 12.4 \\ \bottomrule
\end{tabular}%
}
\caption{Statistics of five Chinese benchmark datasets. We provide the number of sentences, characters, and words. We also compute the out-of-vocabulary (OOV) rate as the percentage of unseen words in the dev and test set.}
\end{table}
}

We employ the CTB5, CTB6, CTB5, and CTB9\footnote{We officially employ the Penn Chinese TreeBank data (LDC2016T13) from the Linguistic Data Consortium. } benchmark datasets from the Penn Chinese Treebank \cite{xue_xia_chiou_palmer_2005}, which has been widely used in research on joint CWS and POS tagging. There are 33 POS tags in CTB. The train/dev/test split for CTB5, CTB6, CTB7 and CTB9 is according to previous studies \cite{zhang-etal-2014-type,yang-xue-2012-chinese,wang-etal-2011-improving,shao-etal-2017-character}. We also employ $\text{UD}$1 and $\text{UD}$2 to denote the datasets using universal tag set and Chinese tag set from UD \cite{nivre-etal-2016-universal}\footnote{We use the UD\_Chinese-GSD dataset with the version 2.4, which extracted from \url{https://universaldependencies.org/}.} following the research of \newcite{tian-etal-2020-joint}, respectively.

\subsection{Implementation}

{
\setlength{\tabcolsep}{2pt}
\begin{table*}[t]
\centering
\resizebox{0.9\textwidth}{!}{%
\begin{tabular}{ccgkgkgkgkgkgk}
\toprule
\multicolumn{2}{c}{\bfseries \textsc{SpanSegTag}} & \multicolumn{2}{c}{\bfseries CTB5} & \multicolumn{2}{c}{\bfseries CTB6} & \multicolumn{2}{c}{\bfseries CTB7} & \multicolumn{2}{c}{\bfseries CTB9} & \multicolumn{2}{c}{\bfseries UD1} & \multicolumn{2}{c}{\bfseries UD2} \\
Encoder & MLP Size & {Seg} & {Tag} & {Seg} & {Tag} & {Seg} & {Tag} & {Seg} & {Tag} & {Seg} & {Tag} & {Seg} & {Tag} \\ \midrule
\multirow{5}{*}{BiLSTM} & 100 & 96.71 & 92.80 & 94.33 & 89.43 & 94.46 & 89.17 & 95.64 & 91.27 & 91.84 & 85.21 & 91.48 & 84.80 \\
 & 200 & 96.90 & 93.08 & 94.90 & 90.06 & 94.70 & 89.36 & 95.96 & 91.57 & 92.36 & 85.92 & 92.27 & 85.78 \\
 & 300 & 97.03 & 93.21 & 95.00 & 90.06 & 94.86 & 89.39 & 96.05 & 91.61 & 92.43 & 86.14 & 92.72 & 85.93 \\ 
 & 400 & 96.82 & 93.27 & 95.18 & 90.16 & 95.04 & 89.53 & 96.15 & 91.54 & 93.02 & 86.45 & 92.84 & 86.03 \\
 & 500 & 97.30 & \bfseries 93.39 & 95.29 & \bfseries 90.19 & 95.10 & \bfseries 89.53 & 96.27 & \bfseries 91.61 & 93.08 & \bfseries 86.74 & 93.12 & \bfseries 86.29\\ \midrule
\multirow{5}{*}{BERT} & 100 & 98.76 & 97.78 & 97.71 & 95.25 & 97.06 & 94.16 & 97.75 & 94.92 & 98.21 & 95.51 & 98.22 & 95.38 \\
 & 200 & 98.78 & 97.71 & 97.66 & 95.25 & 97.11 & 94.24 & 97.78 & 95.07 & 98.23 & 95.64 & 98.21 & \bfseries 95.50 \\
 & 300 & 98.56 & 97.54 & 97.70 & 95.24 & 97.12 & \bfseries 94.27 & 97.74 & 95.02 & 98.35 & \bfseries 95.72 & 98.22 & 95.49 \\
 & 400 & 98.57 & 97.64 & 97.69 & \bfseries 95.26 & 97.05 & 94.18 & 97.80 & \bfseries 95.10 & 98.28 & 95.70 & 98.17 & 95.44 \\ 
 & 500 & 98.81 & \bfseries 97.78 & 97.69 & 95.23 & 97.10 & 94.22 & 97.80 & 95.01 & 98.30 & 95.66 & 98.30 & 95.44 \\ \bottomrule
\end{tabular}%
}
\caption{\label{devtable}Experimental results on development sets of six Chinese benchmark datasets.}
\end{table*}
}

The number of layers of BiLSTM is 1, and the hidden state size of BiLSTM is 200. The dropout rate for embedding, BiLSTM, and MLPs is 0.1. We inherit hyper-parameters from the work of \cite{dozat2017deep}. We trained all models up to 100 with the early stopping strategy with patience epochs of $20$. We used AdamW optimizer \cite{loshchilov2019decoupled} with the default configuration and learning rate of $\text{10}^{\text{-3}}$. The batch size for training and evaluating is up to 5000.

We did fine-tuning experiments based on BERT \cite{devlin-etal-2019-bert}. We trained all models up to 100 with the early stopping strategy with patience epochs of 15 following \newcite{tian-etal-2020-joint}. The dropout rate for MLPs is 0.1. We used AdamW optimizer \cite{loshchilov2019decoupled} with the default configuration and learning rate of $\text{10}^{\text{-5}}$. The batch size for training is 16.

All models were selected based on the performance of the development set. The measure we use for the main result is F-score following previous research. To evaluate F-score of joint CWS and POS tagging, we use the library\footnote{\url{https://github.com/chakki-works/seqeval}.} following the research of \newcite{tian-etal-2020-joint}. We also use paired t-test following the guide of the research \cite{P18-1128} to test the significance of our research.

\subsection{Development Performance}
In Table~\ref{devtable}, we show the performance of \textsc{SpanSegTag} with the output size of MLPs mentioned in subsection~\ref{scoring}. Concerning the BiLSTM encoder, the larger MLP size gives the higher performance in all datasets. Because we regard the joint CWS and POS tagging as a span labeling task, it requires more contextual information. In view of dependency parsing, \newcite{dozat2017deep} chose the MLP size to be 500 for unlabeled parsing. Regarding the BERT encoder, the results of different MLP sizes are not clearly distinguished as those of the BiLSTM encoder since the BERT encoder provides better contextual information. 

{
\setlength{\tabcolsep}{2pt}
\begin{table*}[!htp]
\resizebox{\textwidth}{!}{
\begin{tabular}{lgkgkgkgkgkgk}
\toprule
\multirow{2}{*}{} & \multicolumn{2}{c}{{\bfseries CTB5}} & \multicolumn{2}{c}{{\bfseries CTB6}} & \multicolumn{2}{c}{{\bfseries CTB7}} & \multicolumn{2}{c}{{\bfseries CTB9}} & \multicolumn{2}{c}{{\bfseries UD1}} & \multicolumn{2}{c}{{\bfseries UD2}} \\
 & {Seg} & {Tag} & {Seg} & {Tag} & {Seg} & {Tag} & {Seg} & {Tag} & {Seg} & {Tag} & {Seg} & {Tag} \\
 \midrule
\newcite{jiang-etal-2008-cascaded} & 97.85 & 93.41 & {-} & {-} & {-} & {-} & {-} & {-} & {-} & {-} & {-} & {-} \\
\newcite{kruengkrai-etal-2009-error} & 97.87 & 93.67 & {-} & {-} & {-} & {-} & {-} & {-} & {-} & {-} & {-} & {-} \\
\newcite{sun-2011-stacked} & 98.17 & 94.02 & {-} & {-} & {-} & {-} & {-} & {-} & {-} & {-} & {-} & {-} \\
\newcite{wang-etal-2011-improving} & 98.11 & 94.18 & 95.79 & 91.12 & 95.65 & 90.46 & {-} & {-} & {-} & {-} & {-} & {-} \\
\newcite{shen-etal-2014-chinese} & 98.03 & 93.80 & {-} & {-} & {-} & {-} & {-} & {-} & {-} & {-} & {-} & {-} \\
\newcite{kurita-etal-2017-neural} & 98.41 & 94.84 & {-} & {-} & 96.23 & 91.25 & {-} & {-} & {-} & {-} & {-} & {-} \\
\newcite{shao-etal-2017-character} & 98.02 & 94.38 & {-} & {-} & {-} & {-} & 96.67 & 92.34 & 95.16 & 89.75 & 95.09 & 89.42 \\
\newcite{Zhang2018ASA} & 98.50 & 94.95 & 96.36 & 92.51 & 96.25 & 91.87 & {-} & {-} & {-} & {-} & {-} & {-} \\
\newcite{tian-etal-2020-joint} (BERT) & 98.77 & 96.77 & 97.39 & 94.99 & \bfseries 97.32 & 94.28 & 97.75 & 94.87 & 98.32 & 95.60 & 98.33 & 95.46 \\
\newcite{tian-etal-2020-joint} (ZEN) & \bfseries 98.81 & \bfseries 96.92 & 97.47 & 95.02 & 97.31 & 94.32 & 97.77 & 94.88 & \bfseries 98.33 & \bfseries 95.69 & \bfseries 98.18 & 95.49 \\
\midrule
\textsc{SpanSegTag} (BERT) & 98.67 & 96.77 & \bfseries 97.53 & \bfseries 95.04 & 97.30 & \bfseries 94.50\ddgr & \bfseries 97.86 & \bfseries 95.22\ddgr & 98.06 & 95.59 & 98.12 & \bfseries 95.54 \\
\bottomrule
\end{tabular}%
}
\caption{\label{maintable}Experimental results on test sets of six Chinese benchmark datasets. The symbol \ddgr~denotes that the improvement is statistically significant at $p < \text{0.01}$ compared with TwASP\protect\footnotemark (ZEN) \protect\cite{tian-etal-2020-joint} using paired t-test.}
\end{table*}
}
\footnotetext{We downloaded all pre-trained models of \newcite{tian-etal-2020-joint} from their publicly resource \url{https://github.com/SVAIGBA/TwASP}. However, we can not reproduce the result on the UD2 dataset.}

\subsection{Overall Performance}
We run the final testing experiment with the BERT encoder on six datasets compared to previous results, as shown in Table~\ref{maintable}. Firstly, we can see our \textsc{SpanSegTag} achieve competitive results on CTB5, UD1, and UD2 compared with research of \newcite{tian-etal-2020-joint} using BERT encoder. Our \textsc{SpanSegTag} achieved the competitive or higher F-score on joint CWS and POS tagging even we get the lower CWS performance on CTB5, UD1, and UD2. Besides, our \textsc{SpanSegTag} obtained the higher F-scores of joint CWS and POS tagging on CTB6, CTB7, and CTB9 compared with \cite{tian-etal-2020-joint}.

Compared with \newcite{tian-etal-2020-joint} using ZEN \cite{diao-etal-2020-zen} encoder, we note that the ZEN encoder, which enhances the n-gram information, was better than the BERT encoder on many Chinese NLP tasks \cite{diao-etal-2020-zen}. Though, our \textsc{SpanSegTag} with BERT also obtained the higher joint CWS and POS tagging performance on CTB6, CTB7, CTB9, and UD1. Moreover, our improvements on CTB7 and CTB9 is statistically significant at $p < 0.01$ using paired t-test. We can explain our improvements by modeling all n-grams in the input sentence directly to the word segmentation and POS tagging task via span labeling rather than indirectly according to the work of \newcite{tian-etal-2020-joint}. To sum up, our \textsc{SpanSegTag} does not achieve state-of-the-art performance on all six datasets. However, we obtained significant results on two of the largest joint CWS and POS tagging datasets, including CTB7 and CTB9. To explore the pros and cons of our \textsc{SpanSegTag}, we provide analysis on the section~\ref{analysis}.

\section{Analysis}
\label{analysis}

\subsection{Recall of Out-of-vocabulary and in-vocabulary Words}
\label{oovl}
{
\setlength{\tabcolsep}{2pt}
\begin{table}[http]
\resizebox{\columnwidth}{!}{%
\begin{tabular}{lcgkcgk}
\toprule
\multicolumn{1}{c}{\multirow{2}{*}{}} & \multicolumn{3}{c}{\textbf{$\text{R}_{\text{POS-OOV}}$}} & \multicolumn{3}{c}{\textbf{$\text{R}_{\text{POS-iV}}$}} \\ \cmidrule{2-7} 
\multicolumn{1}{c}{} & \begin{tabular}[c]{@{}c@{}}TwASP\\ (BERT)\end{tabular} & \begin{tabular}[c]{@{}c@{}}TwASP\\ (ZEN)\end{tabular} & \begin{tabular}[c]{@{}c@{}}Our\\ (BERT)\end{tabular} & \begin{tabular}[c]{@{}c@{}}TwASP\\ (BERT)\end{tabular} & \begin{tabular}[c]{@{}c@{}}TwASP\\ (ZEN)\end{tabular} & \begin{tabular}[c]{@{}c@{}}Our\\ (BERT)\end{tabular} \\ \midrule
\textbf{CTB5} & \bfseries 83.81 & \bfseries 83.81 & 82.73 & 97.54 & \bfseries 97.55 & 97.54 \\
\textbf{CTB6} & 83.10 & \bfseries 84.22 & 82.69 & 95.48 & 95.66 & \bfseries 95.68 \\
\textbf{CTB7} & 79.94 & 79.39 & \bfseries 80.19 & 95.20 & 95.25 & \bfseries 95.33 \\
\textbf{CTB9} & \bfseries 79.93 & 78.80 & 78.52 & 95.49 & 95.44 & \bfseries 95.80 \\
\textbf{UD1} & \bfseries 88.67 & 87.40 & 86.13 & 96.64 & \bfseries 96.92 & 96.85 \\ \bottomrule
\end{tabular}%
}
\caption{\label{ooviv}Recall of out-of-vocabulary words and their POS tags ($\text{R}_{\text{POS-OOV}}$) and recall of in-vocabulary words and their POS tags ($\text{R}_{\text{POS-iV}}$). Notably, we do not provide scores on UD2 dataset since we can not reproduce result from the pre-trained model of \protect\newcite{tian-etal-2020-joint}.}
\end{table}

}
Inspired by the research of \newcite{gao-etal-2005-chinese}, we test the performance of detecting unknown words with POS tags ($\text{R}_{\text{POS-OOV}}$) and the performance of resolving ambiguities in word segmentation with POS tags ($\text{R}_{\text{POS-iv}}$), as shown in Table~\ref{ooviv}. The analysis reveals that our \textsc{SpanSegTag} tends to have the higher $\text{R}_{\text{POS-iv}}$ than $\text{R}_{\text{POS-OOV}}$. This analysis motivates us to research the multi-view model of sequence tagging and span labeling in future work.

\subsection{Combination Ambiguity String Error}

{
\setlength{\tabcolsep}{2pt}

}

In addition to $\text{R}_{\text{POS-iv}}$ in subsection~\ref{oovl}, we also follow \cite{gao-etal-2005-chinese} to analyze combination ambiguity string (CAS) errors, as shown in Table~\ref{caserror}. The CAS detection requires a judgment of the syntactic and semantic sense of the segmentation. Hence, we only use the CAS measure in a pilot study. Inspired by \cite{gao-etal-2005-chinese}, we test on a set of 70 high-frequency CASs of each dataset. The result tells that our \textsc{SpanSegTag} solves CASs slightly better than TwASP \cite{tian-etal-2020-joint} on the CTB6, CTB7 and CTB9 datasets. Hence, this error analysis will motivate the research community to improve the joint CWS and POS tagging task.

\begin{table}[http]
\centering
\resizebox{0.9\columnwidth}{!}{%
\begin{tabular}{lgkgkg}
\toprule
 & \textbf{CTB5} & \textbf{CTB6} & \textbf{CTB7} & \textbf{CTB9} & \textbf{UD1} \\ \midrule
TwASP (BERT) & \bfseries 96.43 & 93.72 & 94.26 & 94.61 & 96.40 \\
TwASP (ZEN) & \bfseries 96.43 & 94.88 & 94.23 & 95.47 & \bfseries 97.30 \\
Our (BERT) & 95.71 & \bfseries 95.30 & \bfseries 94.72 & \bfseries 95.56 & \bfseries 97.30  \\ \bottomrule
\end{tabular}%
}
\caption{\label{caserror}CWS accuracies of TwASP \protect\cite{tian-etal-2020-joint} using BERT and ZEN versus our \textsc{SpanSegTag} on 70 high-frequency two-character CASs.}
\end{table}

\subsection{Model Size and Inference Speed}

{
\setlength{\tabcolsep}{2pt}
\begin{table}[ht]
\centering
\resizebox{0.9\columnwidth}{!}{%
\begin{tabular}{lzrzrz}
\toprule
 & \multicolumn{1}{z}{\textbf{CTB5}} & \multicolumn{1}{c}{\textbf{CTB6}} & \multicolumn{1}{z}{\textbf{CTB7}} & \multicolumn{1}{c}{\textbf{CTB9}} & \multicolumn{1}{z}{\textbf{UD1}} \\ \midrule
TwASP (BERT) & 514 & 699 & 716 & 650 & 435 \\
TwASP (ZEN) & 989 & 1,010 & 1,170 & 1,100 & 909 \\
Our (BERT) & \textbf{433} & \textbf{434} & \textbf{435} & \textbf{441} & \textbf{413} \\ \bottomrule
\end{tabular}%
}
\caption{\label{modelsize} Model sizes (MB) of TwASP \protect\cite{tian-etal-2020-joint} using BERT and ZEN versus our \textsc{SpanSegTag}.}
\end{table}

}

In theory, our \textsc{SpanSegTag} is a $O(n^2)$ algorithm due to computing of all possible span representations, which is equivalent to computing of memory network for context features and corresponding knowledge instances from off-the-shelf toolkits in \cite{tian-etal-2020-joint}. In practice, when use GPU Tesla V100 via Google Colaboratory, the inference speed of our \textsc{SpanSegTag} (BERT) and TwASP (BERT) are 264 and 239 (sentence/second), respectively. We notice that we did not count the time TwASP \cite{tian-etal-2020-joint} consuming by running off-the-shelf toolkits. Table~\ref{modelsize} shows that the parameters of our \textsc{SpanSegTag} are independent of the datasets and significant smaller compared with TwASP \cite{tian-etal-2020-joint}.

\section{Related Work}
The one-step approach for joint CWS and POS tagging was proved better than the two-step one by many prior studies \cite{jiang-etal-2008-cascaded,jiang-etal-2009-automatic,sun-2011-stacked,zeng-etal-2013-graph,zheng-etal-2013-deep,kurita-etal-2017-neural,shao-etal-2017-character,Zhang2018ASA}. Besides, \newcite{tian-etal-2020-joint} confirmed the importance of context features and corresponding knowledge instances from off-the-shelf toolkits. Our work is related to \cite{chen-etal-2016-segmentation} in view of using matrix for CWS and to \cite{sun-tsou-1995-ambiguity,chen-goodman-1996-empirical,li-etal-2003-unsupervised,gao-etal-2005-chinese,Ma_2014,chen-etal-2016-segmentation} concerning dealing with ambiguity for CWS.

\section{Conclusion}
In this paper, we propose a neural approach for joint CWS and POS tagging via span labeling. Our proposed approach uses the biaffine operation over the left and right boundary representations of consecutive characters to model the n-grams. Our experiments show that our BERT-based model \textsc{SpanSegTag} achieved competitive performances on the CTB5, CTB6, and UD, and significant improvements on the CTB7 and CTB9 benchmark datasets compared with the current state-of-the-art method TwASP using BERT and ZEN encoders. Our approach does not use any context features and corresponding knowledge instances from off-the-shelf toolkits and a significantly smaller model than TwASP. However, our \textsc{SpanSegTag} has the disadvantage of the complexity and time running. For future work, we will explore the architecture of the BERT model \cite{devlin-etal-2019-bert} for joint CWS and POS tagging because the primitive of BERT also has the complexity of $O(n^2)$ and the self-attention mechanism over the input sentence may be related to span representation.

\bibliography{main}
\bibliographystyle{acl}
\end{document}